\documentclass[conference]{IEEEtran}
\IEEEoverridecommandlockouts
\usepackage{cite}
\usepackage{amsmath,amssymb,amsfonts}
\usepackage{algorithmic}
\usepackage{graphicx}
\usepackage{textcomp}
\usepackage{xcolor}
\def\BibTeX{{\rm B\kern-.05em{\sc i\kern-.025em b}\kern-.08em
    T\kern-.1667em\lower.7ex\hbox{E}\kern-.125emX}}
\begin{document}

\title{A quantum procedure for map generation}

\author{
\IEEEauthorblockN{James R. Wootton}
\IEEEauthorblockA{
\textit{IBM Research - Zurich}\\
Switzerland \\
jwo@zurich.ibm.com
}
}

\IEEEpubid{\begin{minipage}{\textwidth}\ \\[12pt]
978-1-7281-4533-4/20/\$31.00 \copyright 2020 IEEE
\end{minipage}}

\maketitle

\begin{abstract}

Quantum computation is an emerging technology that promises a wide range of possible use cases. This promise is primarily based on algorithms that are unlikely to be viable over the coming decade. For near-term applications, quantum software needs to be carefully tailored to the hardware available. In this paper, we begin to explore whether near-term quantum computers could provide tools that are useful in the creation and implementation of computer games. The procedural generation of geopolitical maps and their associated history is considered as a motivating example. This is performed by encoding a rudimentary decision making process for the nations within a quantum procedure that is well-suited to near-term devices. Given the novelty of quantum computing within the field of procedural generation, we also provide an introduction to the basic concepts involved.

\end{abstract}


\emph{\copyright 2020 IEEE.  Personal use of this material is permitted.  Permission from IEEE must be obtained for all other uses, in any current or future media, including reprinting/republishing this material for advertising or promotional purposes, creating new collective works, for resale or redistribution to servers or lists, or reuse of any copyrighted component of this work in other works}.

\section{Introduction}

Quantum computers offer a radically new form of hardware, with radically new software to go with it~\cite{benioff,feynman,ike-mike}. Over the last few decades, many algorithms have been developed which offer a provable reduction in computational complexity in comparison with conventional digital computing~\cite{deutsch,montanaro}. This `quantum speedup' will be polynomial for some algorithms, such as those for search problems~\cite{grover}, but super-polynomial or even exponential for others, such as factoring~\cite{shor}. Algorithms with proven quantum speedup address variety of different types of problem~\cite{algorithm-zoo,qiskit-textbook}, including optimization, constraint satisfiability and graph theoretic analysis.

Computer games are an application in which the time taken for computation is a very important consideration. Whether a process can fit into a short enough timescale will decide whether it is considered useful. The benefit of quantum computers will be to take problems whose solution requires an infeasibly long time using conventional hardware, and instead solve them within a reasonable time. This will open up new possibilities for the kind of processes that can be fit into the timescale of a loading screen. When considering what applications quantum computers might have in games, we can therefore look to the tasks that will run between turns, within a loading screen or even during the game development process. Tasks that should run within the timescale of a frame, however, will likely be out of scope for some time.

In particular, we will consider the task of procedural generation. This is an area in which many different kinds of computational method are used, with the aim of generating content that obeys a given set of constraints. These could be practical constraints, such as the solvability of puzzles or the removal of situations that could softlock a game, as well as constraints to impose the aesthetics of the game and required difficulty for the player. Since generalized versions of games can easily run into issues of computational hardness~\cite{aloupis:14}, generative methods must be careful to remain computationally efficient. However, compromises made to achieve this can carry the risk that generated samples have a noticeably similar structure, spoiling the illusion that the procedural generation offers infinite variety~\cite{backus:17}.

Even from a high-level overview of known quantum algorithms~\cite{algorithm-zoo}, it is already possible to get an idea of how quantum computers could help in procedural generation. For example, the fact that quantum computers can reduce computational complexity for constraint satisfiability problems could allow for less structured searches of the possibility space. Also, since many approaches to procedural generation use graph-theoretic methods and analysis, the quantum speedups that have been proven in this area may also provide benefits. However, running quantum algorithms such as these will require scalable and fault-tolerant quantum hardware. This is still some years away, and the exact time and resources required by such devices to solve specific problems will depend greatly on as yet unknown specifications of the hardware~\cite{campbell:19}. It is therefore not possible to be more concrete regarding how fault-tolerant quantum computers might be used for games. However, since the above mentioned speedups demonstrate the potential for usefulness in the long-term, this serves as motivation to explore more concretely how useful quantum hardware might be in the near-term.

Current quantum hardware is still very much in a prototype stage~\cite{quantum-volume}. Until now, hardware limitations have meant that it is far easier to emulate quantum devices on a standard laptop than to actually use the real versions\footnote{The process of running simple quantum programs on conventional computers is usually referred to as `simulation'. However, `simulation' is also used to refer to the very different process of using quantum hardware to reproduce quantum dynamics of interest. To avoid ambiguity, we refer to the former as `emulation' throughout this paper.}. It is only with the current generation of 53 qubit devices from IBM and Google that quantum hardware can only be matched by days of dedicated usage on the world’s largest supercomputer~\cite{google-supremacy,ibm-summit}. We are currently in an era of noisy quantum computers~\cite{nisq}, in which one very much has to design quantum algorithms at the machine code level and tailor the method to the exact device used. The path towards demonstrating an advantage over conventional computation is not so clear in this era, and is very much an active area of research~\cite{ibm-nisq}.

The first example of a concrete and useful quantum advantage will likely occur for problems whose structure closely resembles the physics of the quantum hardware. The easiest task that one might give a quantum computer is to simulate a quantum computer (though one might dispute whether this would really constitute a simulation~\cite{horsman:13}). The next obvious, and more useful, task might then be to simulate other quantum systems of interest~\cite{feynman}. Indeed, quantum chemistry is a major focus for applications research in the era of noisy quantum computers~\cite{kandala:17,ganzhorn:19}. It has also has been shown that even unrelated areas such as natural language processing can be expressed in a suitably quantum-compatible manner~\cite{kon:20}. Similarly, can we move beyond ideas of using quantum to simulate quantum, and instead find other simulations that would suit near-term quantum hardware? In this paper we aim to answer this question in the affirmative, for the case of procedural generation and AI for games.

Before continuing, it is worth making a note on the scope of this work. The development of games using quantum computers can be viewed in parallel to the history of games for conventional computers. This history began in the 1950s, with games such as  \textit{Bertie the Brain},  \textit{Nimrod} and  \textit{OXO}. Each was a computer-based implementation of an existing game, and designed more as a showcase of technology, or for education or research, than to be fun. The 1960s brought the first unique play experiences, starting with 1962’s \textit{Spacewar!}, which allowed a player to take control of a space ship via a simulation of orbital mechanics. It wasn’t until a decade later that the first major commercial success emerged: 1972’s  \textit{Pong}.

For the case of games that use quantum computers, the last few years have reproduced some of the landmarks of the 1950s. Simple games, such as a variant of \textit{Battleships}, have been made to serve as a starting point for those wanting to learn quantum programming~\cite{wootton:battleships}. A game was also developed to provide an interactive way to understand benchmarking of the hardware~\cite{wootton:awesomeness}. Many other games have also been made that use quantum programming principles, but run on an emulator, in order to implement quantum-inspired game mechanics~\cite{wootton:history}. Again, this is typically done for the purposes of education. The aim of this paper is to move quantum games into the 1960s, by beginning to explore how quantum computers might actually be useful and provide unique new opportunities in game design.

\section{Quantum networks}

The field of quantum computing is not yet well-known in the study of games or of procedural generation. As such, it falls upon this paper to give a brief introduction. The aim here is not to give a comprehensive understanding of quantum computers and how they work. Instead the aim is to highlight basic properties, especially those pertinent to the approach that we will take.

\subsection{The Qubit}

Conventional digital computing is based on the notion of the `bit': a unit of information that can take one of two possible states. Typically we refer to these states as \texttt{0} and \texttt{1}. In standard `classical' information theory, there is no distinction between the internal state of the bit and its state at output. The event of extracting an output is not a defining moment in the life of a bit. A \texttt{0} is simply a \texttt{0}, and a \texttt{1} is simply a \texttt{1}, whether we look at them or not.

Qubits are similarly a basic unit of information that can take one of two possible states. However, they are described by quantum information theory. This means that they are described in the same way as two-level quantum systems, such as the spin of an electron (or a cat in a box). As such, the moment of output is a defining moment, and is represented in quantum programs by the so-called `measurement' gate. It is at measurement that the qubit must decide whether it is a \texttt{0} or a \texttt{1}. Before that, the output is undefined.

This behaviour is often misinterpreted to mean that quantum systems are simply random. However, though randomness can result from quantum systems, it is easier to understand them by focussing on what they are certain about, rather than the uncertainty.

For a bit, there is only a single way that we can extract an output: simply look at whether it is \texttt{0} or \texttt{1}. For a qubit there are an infinite number of different possible methods, with the result depending on which is chosen. A set of three of these different measurements are sufficient to fully understand a qubit. They are known as x, y and z measurements.

Qubits are typically initialized in a state for which the z measurement results in the output \texttt{0} with certainty. There is similarly a state for which a z measurement will always yield the outcome \texttt{1}. However, making an x or y measurements on either of these states would lead to a completely random result. States also exist for which the results of x measurements are certain, with z and y being random, and similarly for y.

This is most easily described by three parameters

$$-1 \leq \left\langle X\right\rangle, \left\langle Y\right\rangle, \left\langle Z\right\rangle \leq 1.$$

These are related to the probabilities of the outcomes for x, y and z measurements, respectively. $\left\langle X\right\rangle=1$ implies that an x measurement will return \texttt{0} with certainty, and $\left\langle X\right\rangle=-1$ implies the same for an output of \texttt{1}. The value $\left\langle X\right\rangle=0$ implies that the outcome of an x measurement is random. The values of  $\left\langle Y\right\rangle$ and  $\left\langle Z\right\rangle $ similarly predict the behaviour of y and z measurements, respectively.

Valid qubit states must obey the constraint

\begin{equation} \label{heisenberg}
\left\langle X\right\rangle^2 + \left\langle Y\right\rangle^2 + \left\langle Z\right\rangle^2 \leq 1.
\end{equation}

This is the source of some of the behaviour described above. Any state that is certain of its output for one type of measurement measurement (e.g. $\left\langle Z\right\rangle = \pm 1$ and therefore $\left\langle Z\right\rangle^2 = 1$) implies randomness for the other two ($\left\langle X\right\rangle = \left\langle Y\right\rangle = 0$). This constraint is a form of Heisenberg's uncertainty principle.

Note that this constraint can be interpreted as requiring that qubit states must reside on or within a sphere of radius $1$ centred at the origin, with $\left\langle X\right\rangle$, $\left\langle Y\right\rangle$ and $\left\langle Z\right\rangle$ serving as cartesian coordinates. This leads to a popular visualization of qubit states, known as the Bloch sphere.

\begin{figure}[htbp]
\begin{center}
\includegraphics[width=0.8\columnwidth]{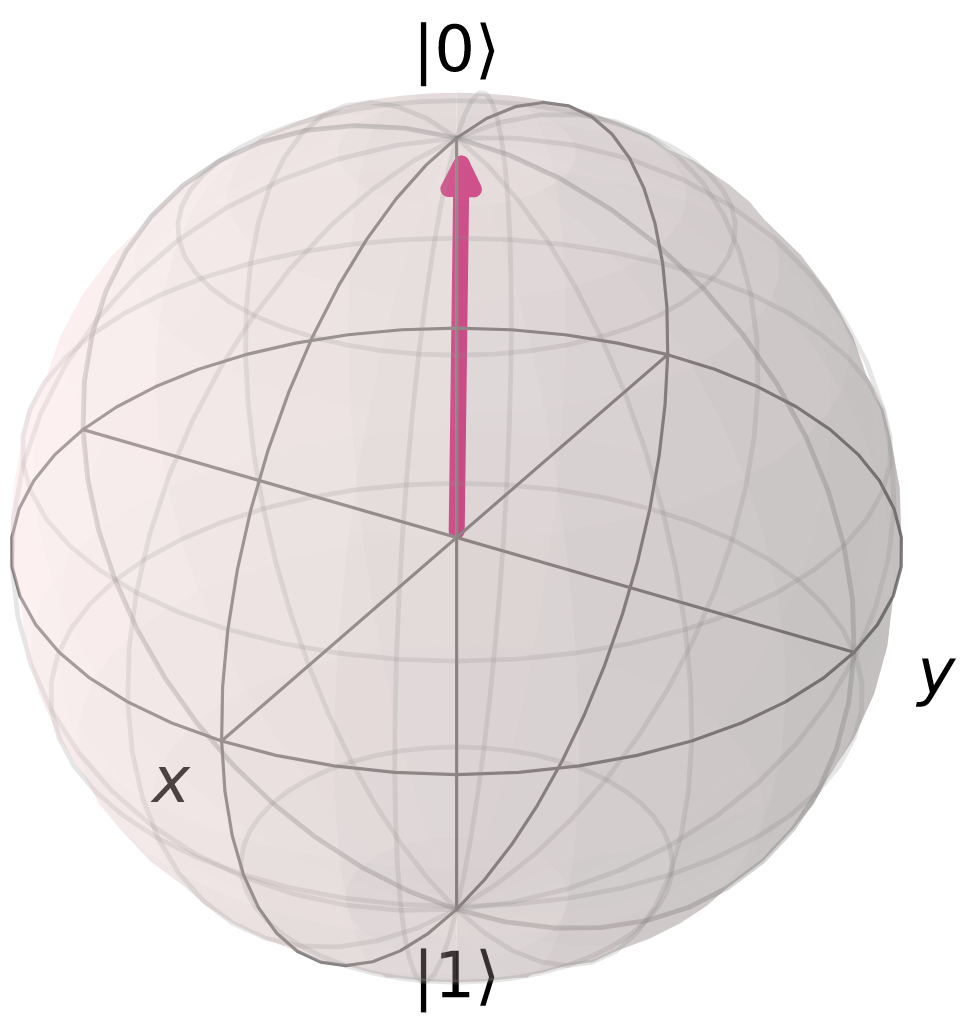}
\caption{Representation of the $\left\langle Z\right\rangle = 1$ state, known as $\left| 0 \right\rangle$, using the Bloch sphere. }
\label{bloch}
\end{center}
\end{figure}

The Bloch sphere is also useful for understanding the gates that can be used to manipulate a single qubit. Each can be represented as a rotation of this sphere, around a given axis by a given angle. As such, note that the value of $\left\langle X\right\rangle^2 + \left\langle Y\right\rangle^2 + \left\langle Z\right\rangle^2$ is invariant under single qubit gates. Changing this value (i.e. the distance from the origin) can only be done with the help of two-qubit or other multi-qubit gates.

\subsection{Two Qubits}

To describe a two qubit state, we need more than just the $\left\langle X\right\rangle$, $\left\langle Y\right\rangle$ and $\left\langle Z\right\rangle$  values of each qubit. We require additional variables which will keep track of correlations between the outputs of the two qubits. For example, $-1 \leq \left\langle X_j X_k \right\rangle \leq 1$, describes the correlations that would arise between the outputs if an x measurement were made of both qubits: $\left\langle X_j X_k\right\rangle = 1$ implies that the qubits will output the same value, $\left\langle X_j X_k\right\rangle = -1$ implies that they output different values and $\left\langle X_j X_k\right\rangle = 0$ would mean that any agreements or disagreements are random. There are similarly $\left\langle Y_j Y_k\right\rangle$ and $\left\langle Z_j Z_k\right\rangle$ to describe correlations between y and z measurement outcomes, $\left\langle X_j Y_k \right\rangle$ for describing correlations for an x measurement of one qubit and a y for the other, and so on.

With these new variables we obtain further constraints on valid states, for example,

\begin{equation} \label{heisenberg2}
\left\langle X_j P_k \right\rangle^2 + \left\langle Y_j P_k \right\rangle^2 \left\langle Z_j P_k \right\rangle^2 \leq 1, P \in \{X,Y,Z\}.
\end{equation}

The correlations in these relations can similarly be visualized as a sphere for which single qubit gates act as rotations.

Given only single qubit operations, only limited correlations can be obtained. For example, two qubits are typically initialized in the state with $\left\langle Z\right\rangle =1$which implies $\left\langle ZZ\right\rangle =1$, but all other correlations are zero. Single qubit operations can change which measurement types experience this correlation, but it remains basically the same.

Sophisticated manipulation of correlations is done via two-qubit operations. We will not describe these in detail here. Instead, as an example of their effect, note that it is possible to use them to create states such as  $\left\langle ZZ\right\rangle =\left\langle XX\right\rangle =-\left\langle YY\right\rangle = 1$. Here perfect correlation (or anticorrelation) is obtained for outputs from each type of measurement, when both qubits are measured in the same way. This comes at the cost of the having $\left\langle X\right\rangle=\left\langle Y\right\rangle=\left\langle Z\right\rangle =0$ for each qubit, which means that each qubit appears completely random when considered on its own. In this case, all the certainty associated with the qubits is directed towards their collective effects.

The state described above is an example of what is known as an entangled state. In fact, it is an example of a maximally entangled state. Entanglement exhibits a property known as monogamy, which means that maximal entanglement between two qubits prevents either from being in any way correlated with any other qubit. More generally, the degree of entanglement between any two qubits limits the degree to with either can entangle with others.

This all serves as an example of how qubits can be thought of as objects described by a set of variables, evolving in a way constrained by a set of constraints.

\subsection{Many Qubits}

We have no need in this paper to introduce the general form for understanding multi-qubit states~\cite{qiskit-textbook}. Suffice it to say that more variables are required to keep track of all possible multi-partite correlations, for any possible subset of qubits and combination of measurements. Specifically, for $n$ qubits the number of these variables scales as $4^n$. The number of constraints on these variables also scales exponentially, but not enough to restrain the exponential number of free parameters needed to fully describe an $n$-qubit system. Indeed, this is the reason why the emulation of quantum software becomes intractable, and hence quantum hardware will be needed.

\begin{figure}[htbp]
\begin{center}
\includegraphics[width=0.95\columnwidth]{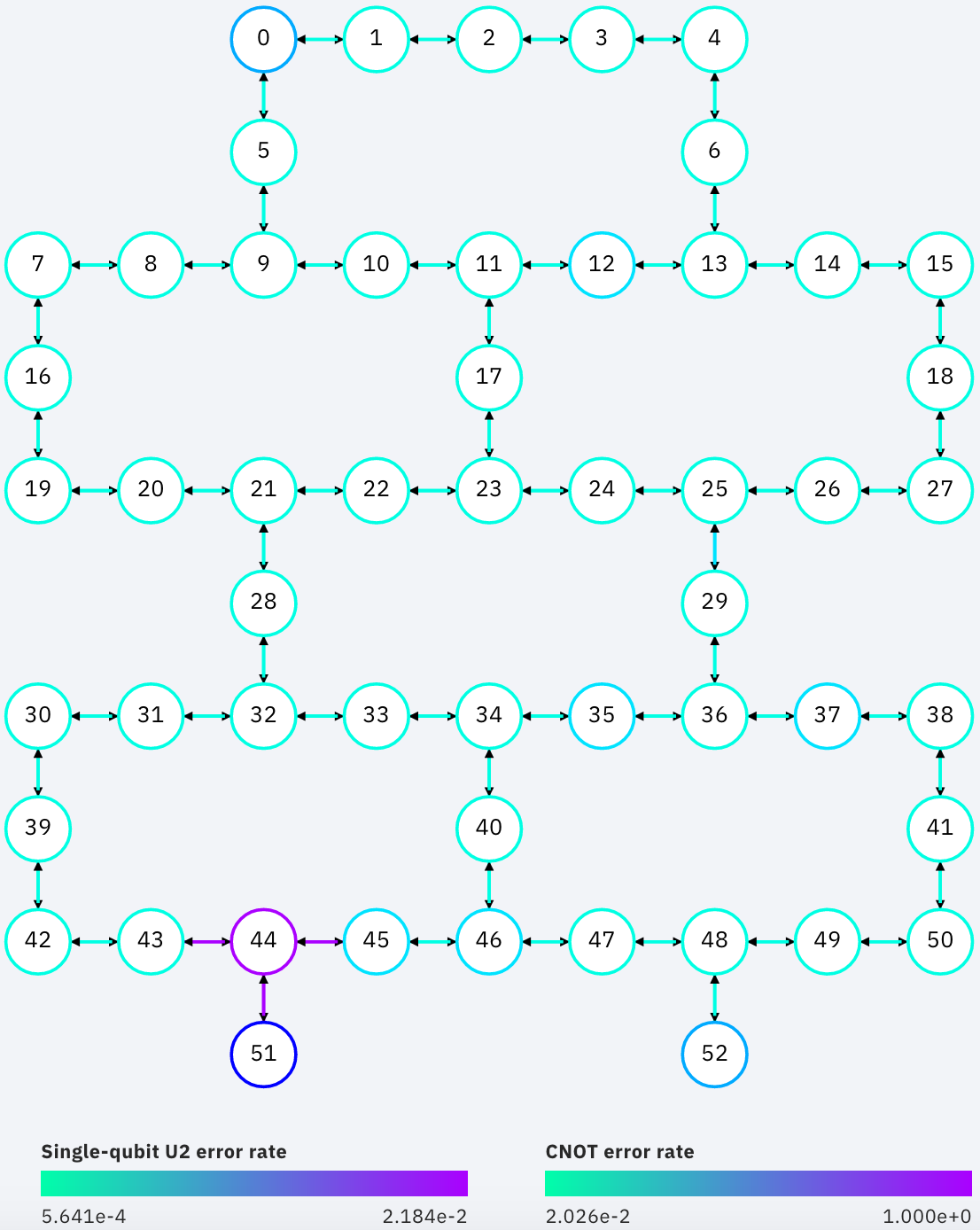}
\caption{Coupling map of the 53 qubit \textit{ibmq\_rochester} device. The numbering convention for qubits, and error rates for single and two-qubit (CNOT) operations are also shown.}
\label{rochester}
\end{center}
\end{figure}

One aspect of multi-qubit systems that will be very important for this work is the availability of two-qubit gates. This depends on the exact form of hardware used to realize the qubits. For the superconducting qubit devices that are currently most prevalent, qubits are located on a two-dimensional chip~\cite{murali:19}. Two-qubit gates are only possible between neighbouring qubits on this chip, and even then only for certain neighbouring pairs. This constitutes the so-called `coupling map' of the device, such as that shown in Fig. (\ref{rochester}), which can differ strongly between differently designed chips.

Any coupling map corresponding to a connected graph can simulate any other without a high cost in terms of computational complexity. However, one must account for the imperfections present in near-term devices and the errors they can cause~\cite{mckay:19}. The additional effort required to simulate alternate coupling maps will lead to the introduction of too many imperfections, causing the output to be dominated by errors. In the near-term, we should therefore stick with the coupling map we are given as much as possible in order to ensure useful results.

\section{Graph-based procedural generation}

To summarize the last section, near-term quantum computers can be well-described as a set of quantum objects (the qubits), placed within a network (as specified by the coupling map), interacting with their neighbours (by means of gates). It is within this structure that we must seek to encode our problems if we wish to solve them on a quantum computer. In the long-term, the encoding can be high-level and abstract. In the near-term, however, imperfections in the devices mean that the more direct the encoding, the better the answer that we can obtain before the effects of errors become dominant.

If our problem is to simulate the interactions of quantum objects within a network, the encoding is clearly extremely direct. This is therefore the problem we will consider, and seek to find applications in procedural generation.

Specifically, we will look towards the use of graphs in procedural generation using a simple version of the method of ~\cite{kybartas:14,kybartas:17}. For example, for narrative generation one can use a graph to encode properties of a set of characters (the nodes) and their relationships (the edges). By looking at the current state of the graph we can see what should happen next: if two characters are found to hate each other, for example, perhaps a fight will result. After this event has taken place, the graph is updated based on the consequences. The next event can then be determined based on the new state of the graph. The evolution of the graph state therefore provides a resource for generating dynamic narratives.

When altering the state of such a graph, one must be careful to obey any consistency relations between different characteristics. For example, suppose we recorded both how friendly and how violent a character was. These are independent enough that they need to be recorded by separate variables, but they are not completely independent: it would seem contradictory to have a character that was both completely violent and completely friendly. As such, the evolution of the variables must be done in such a way that consistency is maintained.

There are many ways that this can be done. One way might be to use a network of qubits. The $\left\langle X\right\rangle$, $\left\langle Y\right\rangle$ and $\left\langle Z\right\rangle$  values of each qubit could be used to encode three characteristics of a character, which are defined such that the condition of Eq. \ref{heisenberg} perfectly sums up their consistency relations, and single qubit quantum gates perfectly describe how their characteristics might evolve. Similarly the relationships should be defined such that they are perfectly captured by variables such as $\left\langle XX \right\rangle$ and their consistency relations, and two qubit gates perfectly describe their evolutions.

The suggestion of the last paragraph is, admittedly, quite speculative. It is not expected that the reader's mind is now abuzz with all the ways that quantum networks would be perfectly suited to their own projects. Instead, in the rest of the paper we will explore a concrete example, to serve as a starting point for exploring how quantum networks could be used in procedural generation.

\section{A theory of quantum government}

We will consider an example in which $n$ qubits are used to simulate the actions and interactions of $n$ entities, which we will think of as the governments of nations. In this way, the quantum network serves as a rudimentary AI. The nations are modelled very simply as having three possible kinds of action: offensive, defensive and explorative. The former two are directed at a particular neighbouring nation. In the next section we will embed these nations into a context in which these actions and their consequences are more concretely defined. However, here we consider them at an abstract level and focus instead on the implementation of the AI.

For each qubit, the $\left\langle X\right\rangle$ variable is used to encode the tendency to defend. Though the full range of this variable is $-1 \leq \left\langle X\right\rangle \leq 1$, only $0 \leq \left\langle X\right\rangle \leq 1$ will be used. Similarly $\left\langle Z\right\rangle$ encodes the tendency to aggression, and $\left\langle Y\right\rangle$ is that to explore.

Each nation will have limited capability to enact its policies. It can fully dedicate itself to one, or have some compromised sharing of priorities. As such it does not make sense for two of these variables to take the maximum value at once, and it makes even less sense for all three to do so. This limit will be naturally implemented via Eq. \ref{heisenberg}.

The two qubits variables describe correlations of policy between two particular nations. For example, while the $\left\langle X\right\rangle$ of each describes a generic tendency to defend, $\left\langle XX\right\rangle$ describes their specific tendency to defend against each other, and so on. The limit of the capability to enact policy against any particular neighbour will be naturally implemented via  Eq. \ref{heisenberg2}. Similarly, the limited degree to which a nation can focus policy on any particular neighbour will be governed by the monogamy of entanglement.

For each nation, $j$, neighbour, $k$, and tactic $P  \in \{X,Y,Z\}$ we calculate the following quantities.
\begin{equation} \label{expectation}
E_{j,k} (P) = \left\langle P_j \right\rangle + \sum_{Q_k \in \{X,Y,Z\} } \left\langle P_j Q_k\right\rangle.
\end{equation}
In this case, the neighbours are defined by the geopolitics of the nations rather than the coupling map of their qubits.

With these quantities for a given nation, we then determine which is maximum. This is then the action that we can expect the nation to employ. Note that this calculation is done as if exploration is performed with respect to a certain neighbour, when in fact it is not.

Calculating these quantities requires us to have access to the necessary single- and two- qubit quantities. However, as mentioned earlier, the output of qubits is in the form of bit values. The quantities are therefore calculated via a tomography process, in which samples are obtained from many runs, which differ only in the form of measurement made at the end. Tomography that covers all possible pairs of qubits can be done with low overhead~\cite{garciaperez:19}. We will therefore do this, to ensure that we can always access the quantities of Eq. \ref{expectation} between nations that share a land border.

Once the nations have taken action against each other, the consequences should result in a change to their states. This can be done by means of single qubit operations. However, clearly the interpretation of these operations as rotations is not intuitive in this case: a nation does not typically respond to a crisis by performing a certain rotation around their aggression axis. Instead it is more natural to think of policy as evolving from its current state to (or towards) another.

This is the way that we will implement single qubit operations on our nations. Tomography tells us a nation's current state, and the actions taken by and against it determine the state to which it is moving (such as the fully defensive $\left\langle Y\right\rangle=1$ state for a nation under attack). The single qubit gate to achieve this is then determined. This gate could be applied itself, or a partial version could be applied instead to take the state a specified fraction of the way to the target.

Note that, as mentioned earlier, single qubit operations cannot vary the value of $ \left\langle X\right\rangle^2 + \left\langle Y\right\rangle^2+ \left\langle Z\right\rangle^2$. More accurately, the target state will then be the closest possible state given this restriction.

Ideally, two qubit manipulations would act in a similar fashion. For example, we might wish to specify the target $\left\langle X_j Z_k\right\rangle=\left\langle Z_j X_k \right\rangle=1$, to implement behaviour consistent with a declaration of war between two nations. Using this target, as well as tomography of the current state, we can then determine the two-qubit gate that gets as close as possible, while otherwise preserving the state as much as possible.

Early tests of this method showed that statistical noise in the tomography made it hard to get the correct gate. Further development is therefore deferred to future work. Instead we will use a method that will indeed get as close as possible to the target, but not be able to otherwise preserve the state as much as possible. We will focus specifically on the $\left\langle X_j Z_k\right\rangle=\left\langle Z_j X_k\right\rangle=1$ case in this work, but the method can be generalized.

The state that fully achieves $\left\langle X_j Z_k \right\rangle=\left\langle Z_j X_k\right\rangle=1$ is uniquely defined (and is a maximally entangled state). The standard way to create it is to take the $\left\langle X\right\rangle=1$ state on both qubits and apply the so called \texttt{cz} two-qubit gate. Our approach will therefore be to apply single qubit gates to both qubits, rotating each as close as possible to their $\left\langle X\right\rangle=1$ state. A \texttt{cz} is then applied.

\section{Procedural map generation}

Using the AI system described in the previous section, we can implement a geopoltical map and history generator which simulates the growth and interactions between nations.

We will consider a simple mechanic as an example, in which everything is driven by the placement of cities. Cities exert influence on their surrounding area, with the strength of the influence at any point depending on the Euclidean distance $d$ from the city, as well as a global parameter, $r$. The influence is chosen to be $1+\min(1,d^{-1})$ for $d \leq r$, and, $d^{-1}$ for $ r< d \leq 2r$, and zero for $d>2r$. The ownership of each position on the map is determined by whether any nation exerts influence and, if multiple do, which exerts the most influence. If a position occupied by the city of one nation comes to have greater influence exerted by another nation, the city will change hands.

Given this mechanic, cities and territories can be gained and lost. It therefore allows us to define the notion of defensive, aggressive and explorative tactics when placing cities. A nation can place cities to defend against a given neighbour by choosing their own point of least influence along the shared border. City placement for exploration is the same, but for the border with unclaimed territory. Aggressive placement against a neighbour is done by finding the border point at which the neighbour's influence is least. Additionally in this case, the influence of the nation itself must be below a certain level to prevent its own cities being placed too close to each other. Specifically, we choose to restrict aggressively placed cities to points at which the influence is no greater than $1+r/2$ (the equivalent from a single city at distance $r/2$).

Nations are restricted in the number of cities that they may have. Specifically, the maximum number is $A/(\pi r^2)$, where $A$ is the number of points within the nation's territory. If a nation deems it necessary to place a further city once this limit is reached, an existing city must be razed. In this case, the city at the position of maximum influence (other than the capital) is chosen. No city can be built on the ruins in future.

Each nation is initially set in a state that is evenly distributed between exploration, attack and defence. Information regarding gains and loss of territory is fed back into the AI to evolve this state. The way this is done depends on the length of the border with unexplored territory, $f$, the area lost, $A_{l}$, and the area gained from other nations, $A_{g}$. Whichever of these is greater for each nation determines whether its qubit is moved toward the fully explorative, defensive or aggressive state, respectively. In the latter two cases, the fraction of the rotation towards this state that is performed is the ratio of the lost or gained area with $\pi r^2$. In the former, the fraction is always $1/4$. 

For the transfer of a city, the effect on the AI is performed using the two-qubit gate described in the last section. This should increase the value of $\left\langle XZ\right\rangle$ and $\left\langle ZX\right\rangle$, and hence increase the likelyhood of specific attack/defense actions between the two nations involved. These operations are applied only when the nations are connected on the coupling map. Otherwise, the losing nation is simply placed into the $\left\langle X\right\rangle=1$ state using a single qubit operation.

With these rules for city placement and the use of the AI, the procedure for map generation is complete. Note that the procedure could also be used to implement a game, simply by handing control over city placement for one or more nations to a player. In this case, the quantum network serves as the AI for the non-player nations, and could provide an advisor for player controlled nations.

\section{Results}

The main test of this procedure will be to verify that the nations seem to modify their tactics based on the situation at hand. This is important whether assessing the procedure as a rudimentary form of AI, or as the the procedural generation of a geopolitical map with an engaging history.

To do this, we can seek to pit our AIs against opponents that do not have these desired characteristics, and determine which is dominant. However, we will need to account for the fact that each individual nation does not have its own individual AI. Instead, they are all collectively simulated by the quantum process. The opponent AIs must therefore be governed by the process also.

To implement this, we simply implement the same process for the opponents, except that the single qubit gates are not applied. Instead of having their state adapt to their changing borders, the opponent nations will simply maintain their initial state. Since this is evenly distributed between exploration, attack and defence, the moves will essentially be chosen randomly. The exception is for two qubit operations, which are applied as normal. This is because there is no way to apply these to the standard nations without also applying them to the opponents.

The metric by which the success of a nation can be measured is its total land area. We will therefore compare the standard nations to the opponents by means of the total land area.

Results were taken for a set-up designed to be compatible with the coupling map of both the \textit{ibmq\_rochester} and \textit{ibmq\_cambridge} devices. These are prototype quantum processors created and made available on the cloud by IBM~\cite{ibm-nisq}. \textit{ibmq\_rochester} has 53 qubits while \textit{ibmq\_cambridge} has 28. The coupling map of \textit{ibmq\_cambridge} is identical to that of the first 28 qubits of \textit{ibmq\_rochester}, as defined by the qubit numbering convention. The coupling map of \textit{ibmq\_rochester} is shown in Fig. (\ref{rochester}).

\begin{figure}[htbp]
\begin{center}
\includegraphics[width=0.95\columnwidth]{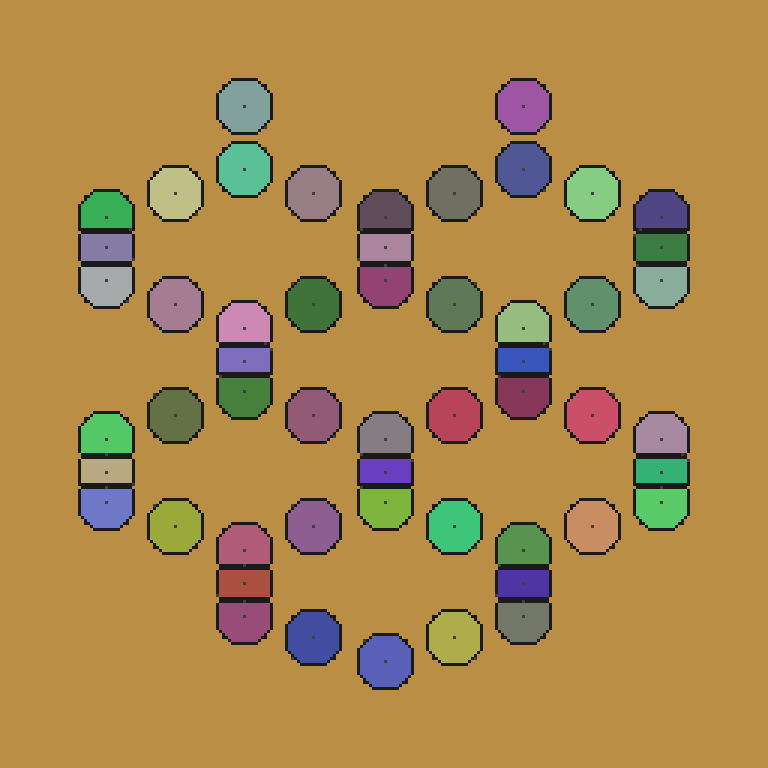}
\caption{Initial city positions for each nation. These correspond largely to the positions of qubits on Rochester (except that it is vertically flipped). However, some deformation of the positions was used such that the Voronoi cells for each point primarily have borders with neighbours on the coupling map.}
\label{initial}
\end{center}
\end{figure}

All runs are done using the same initial city placement, designed such that qubits will mostly tend to share borders with their neighbours on the coupling map. This will help minimize cases for which a city changes hands, but the associated two-qubit gate cannot be applied. The initial layout is shown in Fig. (\ref{initial}).

The total map size used is an $L \times L$ grid for $L=256$. The radius of influence is set to $r=5$. When opponents are present, they are chosen by taking a bicoloring of the nodes of the coupling map. As such, each standard nation will border only opponents, and vice-versa. Equal numbers of samples are taken with the opponents as each of the two bicolorings. Data was taken for small numbers of nations (7, 9 and 11) using emulations of the quantum process: free of any imperfections, and running on a conventional computer. In each case, ten distinct runs were made. Runs were also made on the 28 qubit \textit{ibmq\_cambridge} device and the 53 qubit \textit{ibmq\_rochester}, using all qubits as nations in both cases.

\begin{figure}[htbp]
\begin{center}
\includegraphics[width=0.95\columnwidth]{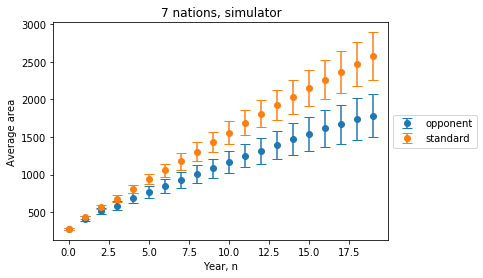}
\includegraphics[width=0.95\columnwidth]{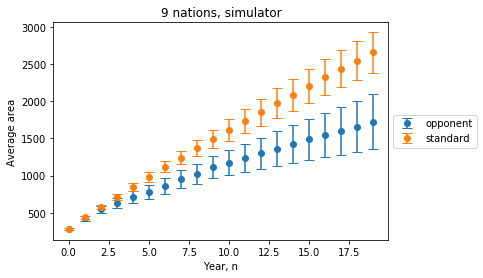}
\includegraphics[width=0.95\columnwidth]{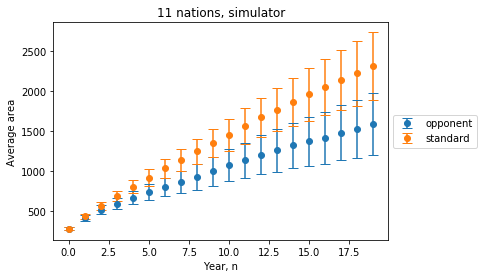}
\includegraphics[width=0.95\columnwidth]{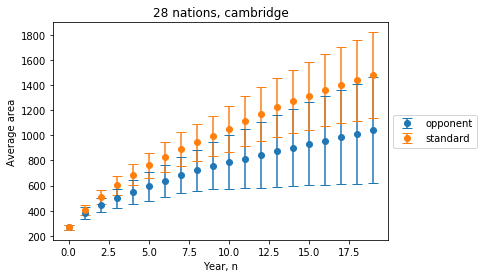}
\includegraphics[width=0.95\columnwidth]{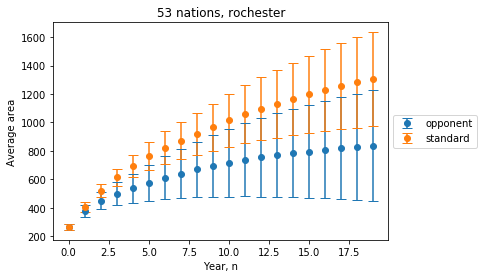}
\caption{Results for emulated runs on 7, 9 and 11 qubits, a 28 qubit run on the \textit{ibmq\_cambridge} device and a 53 qubit run on \textit{ibmq\_rochester}. The mean nation area is shown for each of a sequence of 20 rounds. The length of the bars correspond to the standard deviation of the data at each point.}
\label{default}
\end{center}
\end{figure}

The growth of the mean area shows that the standard nations do indeed have an advantage over their opponents. This demonstrates that the evolution of their quantum state does indeed allow them to adapt to what is happening, and make effective decisions about how to react. However, the standard deviation shows that it is not unlikely for an opponent to be equally successful as a standard nation. Further work will therefore be done to increase the effectiveness of the quantum AI, and increase the gap between it and the opponent.

\begin{figure}[htbp]
\begin{center}
\includegraphics[width=0.95\columnwidth]{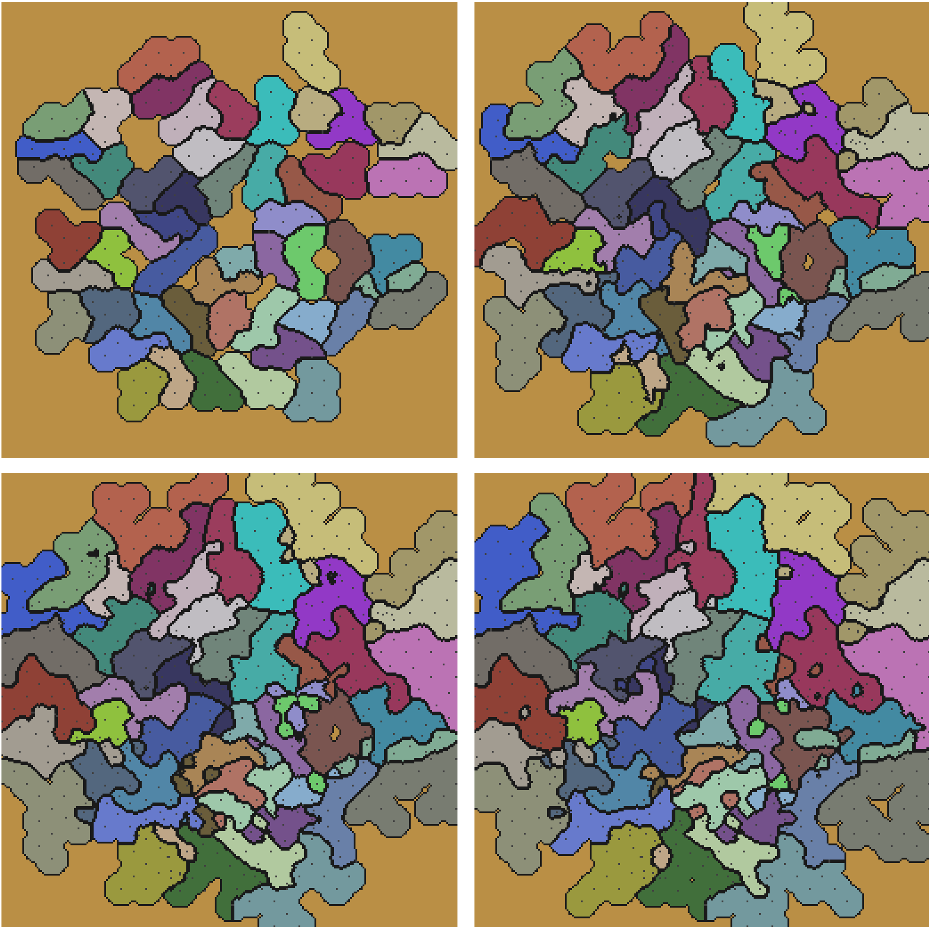}
\caption{Maps generated by a standard run on the \textit{ibmq\_rochester} device after 5, 10, 15 and 20 moves.}
\label{maps}
\end{center}
\end{figure}

Example maps from a standard run, in which all nations are run in the same way, is shown in Fig. (\ref{maps}). These are from runs on the 53 qubit \textit{ibmq\_rochester} device.

Note the regular pattern of city placement is some areas. This is due to the cities being placed for exploration, and hence being put at the points which have minimal influence. In a more sophisticated version, this effect could be mitigated by also accounting for the resource placement and geography of the map. Also note that enclaves and exclaves are relatively common. Despite the fact that they are common also in the real world, these could seem strange to a player. As such a more sophisticated game mechanic which avoids such features will be investigated in further work.

\section{Conclusions}

Quantum computers are a new tool to be explored for many applications, including procedural generation and game design. Here we presented a first step towards this. Specifically, the ability of near-term quantum computers to simulate the state of complex multipartite systems was used to simulate the evolution of a set of nations.

The resulting AI is admittedly rudimentary. If we are to draw a parallel to the history of conventional computers in games, this work has more in common with the IBM's checkers AI of the 1950's that the more recent victory of Watson at \textit{Jeopardy!}. Nevertheless, it is hoped that this work will serve as a motivation for further research and discussion, and that the review of quantum computing provided here might inform similar work.

\section*{Acknowledgment}

Thanks to Matteo Rossi, Guillermo Garcia-Perez, Sabrina Maniscalco, Elsi-Mari Borrelli, and Boris Sokolov for creating the tomography tool used at the Qiskit Camp Europe event, and for discussions thereafter. Thanks also to all on the IBM Quantum hardware development team, and all members of the Qiskit community~\cite{qiskit}.

\bibliographystyle{IEEEtran}
\bibliography{refs}

\end{document}